# A Tool for Spatio-Temporal Analysis of Social Anxiety with Twitter Data

Joohong Lee
Department of Computer Science and Engineering,
Hanyang University
Seoul 04763, Korea
roomylee@hanyang.ac.kr

Dongyoung Sohn
Department of Media and Communication,
Hanyang University
Seoul 04763, Korea
dysohn@hanyang.ac.kr

Yong Suk Choi[*]
Department of Computer Science and Engineering,
Hanyang University
Seoul 04763, Korea
cys@hanyang.ac.kr

## ABSTRACT

In this paper, we present a tool for analyzing spatio-temporal distribution of social anxiety. Twitter, one of the most popular social network services, has been chosen as data source for analysis of social anxiety. Tweets (posted on the Twitter) contain various emotions and thus these individual emotions reflect social atmosphere and public opinion, which are often dependent on spatial and temporal factors. The reason why we choose anxiety among various emotions is that anxiety is very important emotion that is useful for observing and understanding social events of communities. We develop a machine learning based tool to analyze the changes of social atmosphere spatially and temporally. Our tool classifies whether each Tweet contains anxious content or not, and also estimates degree of Tweet anxiety. Furthermore, it also visualizes spatio-temporal distribution of anxiety as a form of web application, which is incorporated with physical map, word cloud, search engine and chart viewer. Our tool is applied to a big tweet data in South Korea to illustrate its usefulness for exploring social atmosphere and public opinion spatio-temporally.

## 1 INTRODUCTION

Anxiety causes a negative effect psychologically and also it can cause various social problems such as low birth rate, suicide-ratio growth, and youth unemployment especially in Korea [1-3]. If the spatial distribution and temporal change of anxiety are well explored and analyzed, then numerous social problems would be coped with. In social sciences, researchers have conventionally tried to grasp for public opinions through formal surveys of samples. However, they often fail to understand actual individual ideas and sentiments about social issues exactly through the surveys [4]. In contrast to the survey approaches, using SNS may be more effective because people tend to post their feelings and opinions to SNS more casually. In our work, we grasp the anxious feeling spread among people from SNS. Since not a few users open spatial information as well as temporal one of their Tweets, we choose the Twitter as data source for analysis of social anxiety.

In order to understand individual moods, a number of lexicon-based approaches [5] analyze the sentiments reflected by Tweets. Other recent researches [6, 7] study how emotions in Tweets are broadcasted spatio-temporally. Such studies are mainly based on some predefined emotion-word-lists and they observe the propagation of general emotions. In this paper, we focus on anxiety that is known to impact individual behaviors in a society, and use a machine learning approach to estimate the degree of anxiety from the Tweets. We also provide a tool to explore and analyze the spatial distribution and temporal change of anxiety.

## 2 NAÏVE BAYES TEXT CLASSIFICATION

### 2.1 Probability Parameters Estimation

In Statistics, a parameter means some value that presents a numerical characteristic of population. It may describe the distribution of random variables assuming that data, observed from population. Among the methods of parameter estimation, we explain two well-known ones, ML (Maximum Likelihood) estimation and MAP (Maximum a Posteriori) estimation.

Naïve Bayes classifier is one of the most popular probabilistic classifiers based on applying Bayes' theorem with Naïve independence assumptions between the features [8, 9]. In many practical applications, parameter estimation for Naïve Bayes models uses the method of maximum likelihood.

$$\hat{c}_{ML} = \underset{c \in C}{\operatorname{argmax}} P(x_1, x_2, \ldots, x_n \mid c)$$
$$= \underset{c \in C}{\operatorname{argmax}} \prod_{i=1}^{n} P(x_i \mid c) \quad (1)$$

$$\hat{c}_{MAP} = \underset{c \in C}{\operatorname{argmax}} P(c \mid x_1, x_2, \ldots, x_n)$$
$$= \underset{c \in C}{\operatorname{argmax}} P(c) P(x_1, x_2, \ldots, x_n \mid c) \quad (2)$$
$$= \underset{c \in C}{\operatorname{argmax}} P(c) \prod_{i=1}^{n} P(c \mid x_i)$$

In text classification, the unknown parameters are classes that should be classified like ML estimation Equation 1 and MAP estimation Equation 2 [9]. $C$ means a set of classes and each class is $c$, and $x$ means words compose a sentence. $\hat{c}$ is the class that makes the equation maximum. Naïve Bayes classifier assumes that all attributes of the examples are independent of each other given the context of the class. Although this assumption is unrealistic in many real-world tasks, Naïve Bayes model is known to often performs classification very well in the text classification.

## 2.2 Analysis of SNS Data with ML Estimation

In this section, we describe the preprocess of SNS data using the morphological analysis and how to classify whether Tweet is anxious or not using the Naïve Bayes classifier.

Morphological analysis is very important process in NLP (Natural Language Processing) to understand linguistic characteristics and structure of words. Tweets comprising a sequence of words are separated as several morphemes, the smallest meaningful units of a language, with POS (Part of Speech) tag such as noun, verb, and adjective. To analyze Tweets, well known Korea morphological analyzer "KOMORAN" is used because we analyze only Korean Tweets [10]. Since our purpose is sentimental analysis, we filter out some insignificant parts of speech. We just use NNG (Noun), VV (Verb), VA (Adjective), MM (Determiner), and MAG (Adverb) as significant POS, that play important roles in speaker's emotion.

A Naïve Bayes classifier is then built using the frequency dictionary of word morphemes that is obtained from the collection of Tweets. Table 1 shows an example template of word morphemes / POS tags and their frequencies that is from the collection of non-anxious Tweets, and Table 2 is from the collection of anxious Tweets. Classification process is explained with some examples. Let Table 1 and Table 2 be the frequency dictionaries that are obtained by tagged Tweets. If there is a sentence consisting of $w_A$, $w_B$, and $w_D$, then by ML estimation, the probability that this sentence is classified into Anxiety is $P(w_A, w_B, w_D \mid Anxiety) = \frac{30}{100} \times \frac{10}{100} \times \frac{20}{100} = 0.006$, and the probability that is classified into Non − Anxiety is $P(w_A, w_B, w_D \mid Non - Anxiety) = \frac{200}{1000} \times \frac{100}{1000} \times \frac{100}{1000} = 0.002$. This sentence can be classified into $Anxiety$ because the probability of Anxiety is greater than the one of Non-Anxiety. If there is a sentence consisting of $w_B$, $w_D$, and $w_F$, then in the same way, the probability of Anxiety is $P(w_B, w_D, w_F \mid Anxiety) = \frac{10}{100} \times \frac{20}{100} \times \frac{10}{100} = 0.002$, and the probability that is classified into Non-Anxiety is $P(w_B, w_D, w_F \mid Non - Anxiety) = \frac{100}{1000} \times \frac{100}{1000} \times \frac{400}{1000} = 0.004$. So, this sentence can be classified into Non-Anxiety.

**Table 1: Frequency dictionary of Non-Anxious Tweets**

| Word / POS | Frequency |
|---|---|
| $w_A$ / NNG | 200 |
| $w_B$ / VV | 100 |
| $w_C$ / VA | 200 |
| $w_D$ / NNG | 100 |
| $w_E$ / VV | 0 |
| $w_F$ / MAG | 400 |
| Total | 1,000 |

**Table 2: Frequency dictionary of Anxious Tweets**

| Word / POS | Frequency |
|---|---|
| $w_A$ / NNG | 30 |
| $w_B$ / VV | 10 |
| $w_C$ / VA | 0 |
| $w_D$ / NNG | 20 |
| $w_E$ / VV | 30 |
| $w_F$ / MAG | 10 |
| Total | 100 |

For another example, if there is a sentence consisting of $w_A$, $w_C$, and $w_E$, then the probabilities of both classes are zero because of $P(w_C \mid Anxiety)$ and $P(w_E \mid Non - Anxiety)$ are all zero. To solve this "zero frequency" problem, we use Add-One Smoothing, also called Laplace smoothing. Add-One Smoothing allows the assignment of non-zero probabilities to the words which do not occur in the collection. For each class, every word is given with extra 1 frequency and thus total frequency is added with the size of word vocabularies. Thus, the example sentence has $P(w_A, w_C, w_E \mid Anxiety) = \frac{30+1}{100+6} \times \frac{0+1}{100+6} \times \frac{30+1}{100+6} \approx 0.000806$, and $P(w_B, w_D, w_F \mid Anxiety) = \frac{200+1}{1000+6} \times \frac{200+1}{1000+6} \times \frac{0+1}{1000+6} \approx 0.000039$. As a result, the sentence consisting of $w_A$, $w_C$, and $w_E$ can be classified into $Anxiety$.

## 4 EXPERIMENTS

### 4.1 Twitter Data Collection

Using a Tweet crawler (of Twitter4j) distributed publicly as Open API [11], we collected 2,711,807 Tweets that contains both spatial and temporal information, from Feb. 2016 to Nov. 2017 in South Korea. Spatial and temporal information comprises time and location (longitude and latitude) of the Tweets. From all the collected Tweets, we random-sampled and then tagged 81,873 Tweets manually with a binary value, whether the Tweet contains anxious expression or not. We categorized anxiety into 18 emotions[1] to reflect comprehensive unpleasant feelings that may be related to social events, issues and atmospheres. For each Tweet, three graduate students (in communication studies) identified the semantical existence of this emotional expressions independently, and then they cross-checked the results. We divided all the tagged Tweets into 75,051 training and 6,822 testing sets. In the training set, the number of anxious Tweets was 7,366 (about 10%).

### 4.2 Comparison of ML and MAP Estimations

Using the testing set of 6,822 tagged Tweets, we evaluate performance of our classifier with ML and MAP estimations.

**Table 3: Performance of ML and MAP Estimations**

| Method | Recall of *Anxiety* | Recall of *Non-Anxiety* | Accuracy |
|---|---|---|---|
| ML | 0.8550 | 0.8292 | 0.8317 |
| MAP | 0.5069 | 0.9641 | 0.9371 |

As shown in Table 3, MAP estimation is better than ML estimation in terms of accuracy. However, MAP shows bad performance in recall of anxiety. Because the training set is greatly biased toward non-anxiety data, $P(Non-Anxiety)$ is much bigger than $P(Anxiety)$ and thus the results of MAP estimation are likely to be mostly $Non-Anxiety$. As a result, MAP estimation shows poor performance in recall of anxiety and good performance in accuracy. In contrast, ML estimation shows generally fair performance, because it is not influenced by the bias of $P(Anxiety)$ or $P(Non-Anxiety)$. In our tool, discovering anxious Tweets well is the key point so we focus on recall of anxiety first and then accuracy. Accordingly, ML estimation is more appropriate than MAP estimation for our purpose. So, we apply ML estimation to Naïve Bayes classifier of our tool [12].

### 4.3 Classification Criterion

Naïve Bayes Classifier applying ML estimation calculates the probabilities of anxiety and non-anxiety for each Tweet. To improve performance of classifier, we formulate a criterion that if Equation 3 is true, then the sentence is classified into $Anxiety$.

$$\frac{P(Tweet\,|\,Anxiety)}{P(Tweet\,|\,Non-Anxiety)} > Threshold \quad (3)$$

Fundamentally, Naïve Bayes classifier based on ML estimation classifies Tweets using this Equation 3 with $Threshold = 1$. However, it may not suit our application very well. So, we also experiment with changing Threshold value. Figure 1 is result of this experiment. We mainly focus on recall of anxiety and accuracy because it is very important to find out the obvious anxious Tweets in our application. Moreover, we use the product of recall of anxiety and accuracy in order to reflect the two measures at the same time. In Figure 1, the product of the two measures has a maximum value 0.7294 at Threshold=2.5, then, we adopt it as the threshold value of Equation 3.

---

[1] 18 emotions are "nervousness", "perplexity", "worry", "excitement", "restlessness", "frustration", "apprehension", "discomfort", "fear", "turmoil", "yearning", "depress", "gloom", "hostility", "desperation", "dismay", "petulance", and "malaise". They can be semantically overlapping each other especially in Korean expressions due to the differences of Korean and English vocabularies.

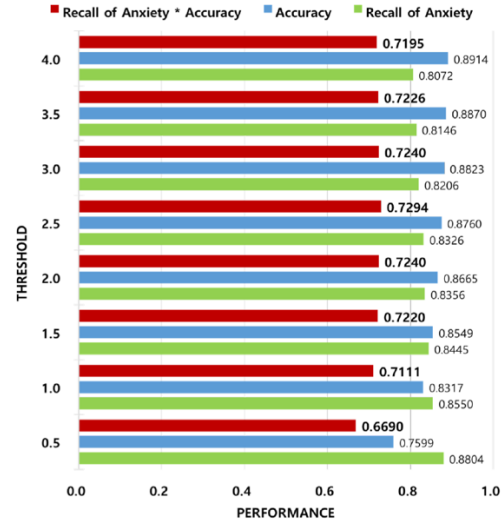

**Figure 1 Performance of Classifier over Threshold Values**

## 5 VISUALIZATION OF ESTIMATED ANXIETY

We show the degree of anxiety based on spatio-temporal information obtained from classified Tweets, and it is visualized as a form of web application[2] shown as Figure 2, which has various functionalities such as map, time controller, word cloud, etc.

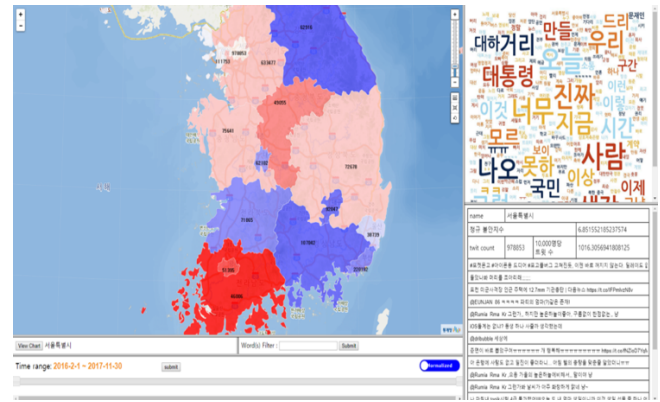

**Figure 2 A Tool Visualizing Spatio-Temporal Social Anxiety in South Korea**

### 5.1 Map: Spatial Viewer

In Figure 2, a map located in the upper left corner is the main function of the tool. Tweets classified by Naïve Bayes classifier are summarized on the map according to their spatial information. The red areas on the map are the relatively anxious region, while blue areas are not. The denser the color and the greater the degree of anxiety or non-anxiety. A high degree of anxiety means occurrence of a lots of anxious Tweets in its region. The map can be zoomed in or out via +/- icons. If the map is zoomed in, each province is divided into counties of smaller unit.

---

[2] Our tool can be accessed at
http://166.104.140.75:62000/Emotional_Analyzation/EA/main.html

## 5.2 Time Controller: Temporal Viewer

Time controller can help manipulating map with regard to temporal information. Time controller places in bottom side, and through dragging scroll-bars at both ends, they can control the range of time. If they click the submit button above the controller, the range of time is applied to map. The time controller may help to explain or predict social events over time.

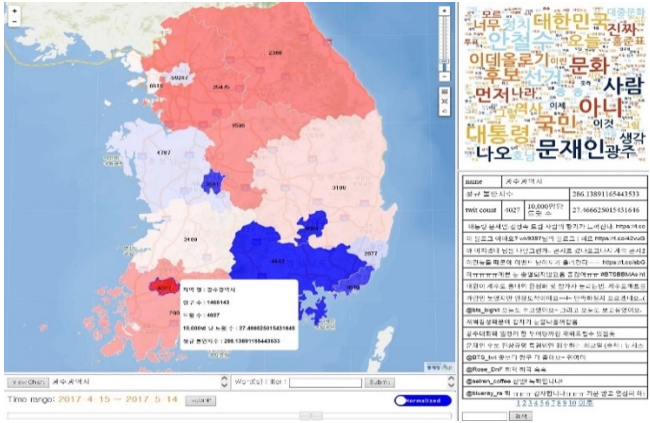

**Figure 3 Distribution of Social Anxiety in South Korea from April 2017 to May 2017**

For example, Figure 3 shows the result of all the Tweets from April 2017 to May 2017, the period just before presidential election, by adjusting time interval. The distribution of anxiety by regions in Figure 3 differs from the one in Figure 2, and especially it seems quite comparable to the regional distribution of actual election results. Thus, political atmosphere and election results might be well explained or predicted by analyzing spatial differences and temporal changes of anxiety with our tool.

## 5.3 Word Cloud and Search Engine

Word cloud and posted Tweet texts of selected region appear at the right side as in Figure 2 and 3. If you click a certain region on the map, then can see all the Tweets posted in that region, and their detailed meta-information like name of city, total counts of Tweets, etc. A word cloud shows hot keywords of selected region and time interval at a glance. A search engine (Words Filter) allows identifying all the Tweets containing the search words submitted.

## 6 CONCLUSIONS AND FUTURE WORK

In this work, we suggest an appropriate probabilistic method to construct a (sub)optimal classifier that estimates social anxiety of Tweets. The results of classification are visualized spatio-temporally in our tool. We expect that spatio-temporal distribution of social anxiety can be effectively analyzed by many researchers in the field of social sciences using our tool. Currently, we are collecting much more training Tweets tagged with more detailed classes to enhance the performance of our tool. We are focusing on balancing the amounts of anxious and non-anxious training Tweets, which will improve our method substantially employing MAP estimation. In the near future, our tool will be extended for other social emotions in order to analyze the relationships between various emotions and social events in more sophisticated ways.

## ACKNOWLEDGMENTS

This research was supported by Basic Science Research Program through the National Research Foundation of Korea(NRF) funded by the Ministry of Education(NRF-2015R1D1A1A01060950) and also supported by the Technology Innovation Programs (No.10077553 and No.10060086) funded by the Ministry of Trade, industry & Energy (MOTIE, Korea)